# A Vehicle Detection Approach using Deep Learning Methodologies

Abdullah Asım YILMAZ[1]
Mehmet Serdar GÜZEL[2]
İman ASKERBEYLİ[3]
Erkan BOSTANCI[4]

[1,2,3,4]Computer Engineering Department
Ankara University
Golbasi 50th Year Campus, I Block, 06830, Ankara,TR
Corresponding author email: [2]mguzel@ankara.edu.tr

*Abstract:* The purpose of this study is to successfully train our vehicle detector using R-CNN, Faster R-CNN deep learning methods on a sample vehicle data sets and to optimize the success rate of the trained detector by providing efficient results for vehicle detection by testing the trained vehicle detector on the test data. The working method consists of six main stages. These are respectively; loading the data set, the design of the convolutional neural network, configuration of training options, training of the Faster R-CNN object detector and evaluation of trained detector. In addition, in the scope of the study, Faster R-CNN, R-CNN deep learning methods were mentioned and experimental analysis comparisons were made with the results obtained from vehicle detection.

## 1 Introduction

Today, many new technological developments have occurred. As a result of these technological developments, people may face several cruical problems. Some of the negativities to be experienced with the detection of such problems can be minimized through various approaches.

Because of this; it is necessary to fall in parallel with technological and scientific developments, traffic accidents are still at the forefront people's daily life both in our country and also all over the world. Compared to air, sea and railway traffic, traffic on highways continues to be a significant problem in current life [1]. Because no human being can not be thought without driving out on a daily basis, traffic accident are  considered to be an ordinary event of each human being. For this reason, it is essential for any person who involved in a traffic accident to objectively present the possible defect situations in a technical, legal and scientific way. So, vehicle detection and tracking systems gather lots of attention from the reasearchers and is highly focused on by scientis. Therefore, there are studies have been carried out about about vehicle detection and tracking systems and day after day new solutions and algorithms are developed with new studies [1].

Deep Learning and relevant techonologies are another mostly elaboreted topic in recent periods, as well. If we define deep learning methods that they are the methods that includes artificial neural networks , comprising at least one hidden layer. They arecapable of performing feature detection process automatically from large amounts of tagged training data. Deep learning methods utilize algorithms known as Neural Networks, which are inspired by information processing methods of biological nervous systems such as  brain and these methods allow computers to learn what each data represents and what each corresponding model actually means.

In this study, vehicle detection and deep learning approaches are combined. Moreover, our vehicle detector on the sample vehicle data sets are individually and successfully trained using fast R-CNN and R-CNN deep learning methods respectively. The trained vehicle detector is tested on the test data and efficient results are obtained from vehicle detection problem. In addition, the success detection rate of the trained detector has been tried to be maximized as much as possible and experimental analysis comparisons are made with the results obtained from the methods.. Overall section will detail the proposed method, wheras will illustrate implemnetaiton details and also the exerimental results. Finally , the study will be concluded in section 4.

## 2  Methods used for Vehicle Detection

The working method consists of six main stages. These are respectively; loading the data set, the design of the convolutional neural network, configuration of training options, training of the Faster R-CNN object detector, evaluation of trained detector. These stages and conventional and the faster R-CNN methods will be discussed in this section.

### 2.1 R-CNN (Regions with Convolutional Neural Network Features)

The R-CNN approach combines two basic concepts. From these, the first is to carry out efficient convolutional neural networks from bottom to up region proposals to locate and dismember objects. Next, when the label training data is insufficient, it is followed by a supervised training for a field-specific fine tuning task, which provides significant performance improvement. The method is named R-CNN (CNN-enabled regions) because Regional proposals are combined with CNNs.

The working object detection system composed of three modules. Firstly, it categorically produces independent region proposals. These essentially describe the candidate detection set that can be used by the detector. The second module includes a convolutional neural network, producing an attribute vector of constant length from every region. The third module, on the other hand, includes a cluster of linear SVMs that are specific to the class used for assortment of regions [8].

#### 2.1.1 Region Proposal

Various recent studies have provided methods to produce categorical independent zone recommendations.These methods have examples such as the objectness of image windows [1] , selective Search for Object Recognition [3] , category independent object proposals [4] , object segmentation using constrained parametric min-cuts [5] , Multiscale combinatorial grouping [6] and so on [7]. These methods establish cells by implementing convolution neural network with  square cuts. Although R-CNN is not based on the specific zone proposal method, R-CNN performs its operations using selective search methods to provide comparison with the predetermined work.

#### 2.1.2 CNN (Convolutional Neural Network) for Feature extraction

In this study, a feature vector of size 4096 were extracted from each region proposal with Caffe deep learning framework. Features were calculated by forwarding the average output 227x227 red-green-blue image with five convolution layers and two completely connected layers.

In order to calculate an attribute in a region proposal, the image data is first converted to a form compatible with CNN. (In this study, fixed entrances of 227 * 227 pixels in size are used.). Then,the most simple of the possible transformations of the random-shaped regions was selected. Here, all the pixels in a tight bounding box around the candidate area are resolved unto the required size, regardless of the size or aspect ratio. Before dissolving, the tight bounding box was expanded to provide w pixels skewed picture content around the box at the skewed dimension (w = 16 was used). In addition, a simple bounding box regression was used to expand the localization performance within the application [13]. This is shown in the following equation (1). The details of this equation can be seen in  [8].

$$w_* = \underset{w_x}{argmin} \sum_i^N (t_*^i - w_*^T \emptyset_5(P^i))^2 + \lambda ||W_*||^2$$

(1)

#### 2.1.3 Classify Regions

In this study, selective search was performed on test images to obtain approximately 2000 region proposals at test time. Each proposal has been resolved and advanced through CNN for the calculation of attributes. Then, for each class, each produced attribute vector is scored using the trained support vector machine (SVM). Considering the scored regions, greedy non-maximum suppression is applied independently when there is high-intersection (IoU) overlap with the selected zone with a higher rating over a learned threshold for a rejected region.

### 2.1.4 R-CNN Training

The details of R-CNN training will be discussed at the following sub-sections.

#### 2.1.4.1 Supervised Pre-training

CNN was previously trained on a large auxiliary data set (ILSVRC2012 classification [9,13]) using only image-level additional tags. CNN was previously trained on data set (ImageNET ILSVRC2012 [9,13]) using only additional tags. This training was carried out using Caffe Deep Learning framework.

#### 2.1.4.2 Domain-Specific Fine-Tuning

In order to adjust Convolutional Neural Network to new task and domain name , SGD training was performed to function parameters using only warped region proposals. Convolutional Neural Networks ImageNetspecific 1000 way classification layer has been changed over with the N+1 way classification layer. Convolutional Neural Network framework has not been changed here. (N = 20 for VOC and N = 200 for ILSVRC2013).

All region proposals, which are equal to or greater than 0.5 iou overlap value, were accepted as positive for the box class and others were accepted as negative. In each SGD iteration, 32 positive windows and 96 background windows are properly sampled to create a mini stack of 128 sizes.

#### 2.1.4.3 Object Category Classifiers

Here, binary classifier training was used to perceive cars. It is a positive example of an image area in which a car is tightly enclosed. In a similar way, a background region that is not interested in cars is a negative example. It is unclear how a partially overlapping region of the car should be labeled. the unclear state is solved by specifying an IoU overlap threshold value. Areas below this threshold value are identified as negative and those above the threshold value as positive.The overlap threshold "0.3" was chosen by conducting a grid search on the verification set. Once the features are removed and the training tags are applied, SVM is applied optimally to all classes.

### 2.2 Faster R-CNN

The Faster R-CNN composed of two component. The first component is a conventional network used to propose zones called RPN, and The second component is the Faster R-CNN detector which utilize the region proposals. The whole system comes from a single composite network created for object detection [10].

The first component is a conventional network used to propose regions called RPN (Region Proposal Network)

#### 2.2.1 Region Proposal Networks (RPN)

In this study, RPN receives as input image and produces a set of rectangle object tender which all have objectivity score. The RPN is designed with a fully convoluted network. Since calculations are shared with a Faster R-CNN object detection network, it is assumed that both networks share a common set of layers of convolution.

A mini network is moving on the exit of the convolution property map by the last shared convolution layer to produce region proposals. As an input, it takes the space window of the convolutional property map $n \times n$(used as n=3). All sliding windows in work are matched to low dimensional property. This feature composed of two sister fully bound box-regression and box-classification layers. In this mini network, all the fully connected layers are shared in all spatial locations.This framewoek is carried out by the convolution Layer and following the two brothers 1 x 1 convolution layer.

#### 2.2.1.1 Training RPNs (Region Proposal Network)

In this study, RPNs are trained end-to-end with backpropagation and SGD. In order to train this network, "image-centric sampling" strategy is applied.In the study, all the batchs come from the images involving negative and positive sample anchors.

When the missing functions of all the anchors are optimized here, the orientation of the negative examples is realized. For this reason, a random sample of 256 anchors is shown in an image instead. According to this, if there are more than 128 positive samples, it is filled with stacked samples. Otherwise, it is filled with negative examples. In addition, following the multitasking loss for Fast R-CNN, the objective function has been reduced to a minimum. This loss function is shown in the following equation (2) , the details of the Equation also can be seen in [10].

$$L(\{p_i\}, \{t_i\}) \frac{1}{N_{cls}} \sum_i L_{cls(P_i, P_i^*)} + \lambda \frac{1}{N_{reg}} \sum_i p_i L_{reg(t_i, t_i^*)}$$

(2)

### 2.2.2 Sharing Features for RPN and Faster R-CNN

Until now, it has been explained how to train a network to generate a region proposal without taking into account the region-based object detection. Here, Faster R-CNN is used for the detection network. After that, a unified network learning algorithm consisting of shared convolution layer RPN and Faster R-CNN is defined.

Here both RPN and Faster R-CNN are trained independently and the layers of convolution are characterized by different forms. For this reason, instead of learning two separate networks, techniques have been developed that are able to share layers of convolution between two networks. There are three techniques to train networks in shared properties. These are alternating training which are, approximate joint and non-approximate joint trainings respectively.

### 2.3 Comparison of Faster R-CNN and R-CNN Methods

Today, the most sophisticated object detection networks are based on region proposal algorithms for the description and identification of object locations.Faster R-CNN has put forward the regional proposal calculation as a bottleneck by reducing the working time of these detection networks in R-CNN. In the Faster R-CNN, a Region Proposition Network (RPN) is implemented that shares full image convolution characteristics using the detection network, so that almost free region proposals can be made. Faster R-CNN, along with the improved RPN, do not require external zone recommendations, unlike R-CNN. In addition, the RPN improves the quality of the district proposal and thus improves the overall accuracy and speed of object detection.

## 3 Detail's of Implementation

This study aims to successfully train the vehicle detector on the sample vehicle data sets using the Faster R-CNN and R-CNN deep learning methods, shown in Section. It also aims to achieve maximal results for vehicle detection by testing the trained vehicle detector on the test data. In addition, the results obtained from these methods are compared with experimental analysis. To do this, the Caffe deep learning framework is used on the Matlab Program.

The purpose of this study is to successfully train our vehicle detector using R-CNN, Faster R-CNN deep learning methods on a sample vehicle data sets and to optimize the success rate of the trained detector by providing efficient results for vehicle detection by testing the trained vehicle detector on the test data. The working method consists of six main stages. These are respectively; loading the data set, the design of the convolutional neural network, configuration of training options, training of the Faster R-CNN object detector, evaluation of trained detector. In addition, in the scope of the study, Faster R-CNN, R-CNN deep learning methods were mentioned and experimental analysis comparisons were made with the results obtained from vehicle detection.

Our application consists of 6 main steps. These; loading the data set, the design of the convolutional neural network, configuration of training options, training of the Faster R-CNN object detector, evaluation of trained detector.

First, the loading of the data set is performed. In this study, two different vehicle dataset were employed. The first dataset includes approximately 350 images [11] and 1000 images are obtained from the second public vehicle dataset [12] . Each image in these datasets includes one or two tagged vehicle samples . In this study, the training data is stored on a table. The existing columns on the table contain the contents of the path of the image files and the ROI tags for the vehicles. In addition, in this section, the data set is divided into training and test sets to train the detector and evaluate the detector. In this part, in order to train the detector, 60% of the data is selected as the training set, and the remaining data is selected and used as the test set for evaluation of the detector's performance. Afterwards, the process of the design of the CNN has been performed. In this phase, the type and size of the input layer are defined. For classification tasks, the input size is chosen as the size of the training images. For detection tasks, CNN should analyze smaller portions of the image, so the input size was chosen as a 32x32 input size, similar to the smallest object in the dataset. Next, the middle layer of the network is defined.. The medium layer created here is made of repeating blocks of convoltional, relu (rectified linear units) and pool layers. Finally, a final layer consisting of fully connected layers and a softmax loss layer was created. Next, the design of the CNN is completed by combining the input, middle and final layers.

The third step is to configuration of training options. At this stage, the training options for the Faster R-CNN method have been configured in four steps. In the first two steps, the region proposals used in the Faster R-CNN and the detection networks are trained. In the last two steps, the first two steps were merged to form a single network. Later, the training options configuration for the R-CNN method were performed in one step. Here, the network training algorithm is configured using an SGDM with an initial learning rate of 0.001.

In the fourth step, training of fast R-CNN and R-CNN object detectors was carried out. At this stage, image patches were extracted by selecting from the training data during the training process. Two kinds of name value pairs are used here. It has been checked which image patch is used with these. Positive training samples here are examples with 0.6 to 1.0 with accuracy boxes, as measured by the bounding box intersection of the unity metric. Negative education examples are examples that overlap between 0 and 0.3. Maximized values for these parameters were selected by testing the trained detector on the verification set.

Finally, the process of evaluation of trained fast R-CNN and R-CNN detectors were carried out. At this stage, the detection results were collected and evaluated by running the trained detectors on the test set.

## 4 Conclusion

The proposed vehicle detector has been successfully trained by using Faster R-CNN and R-CNN deep learning methods on the sample vehicle datasets and the vehicle detection process has been successfully performed by the trained vehicle detector being tested on the test data set. As the output of the study, the image frames are shown in Figure 1 and Figure 5 for fast R-CNN and in Figure 2 and Figure 6 for R-CNN.

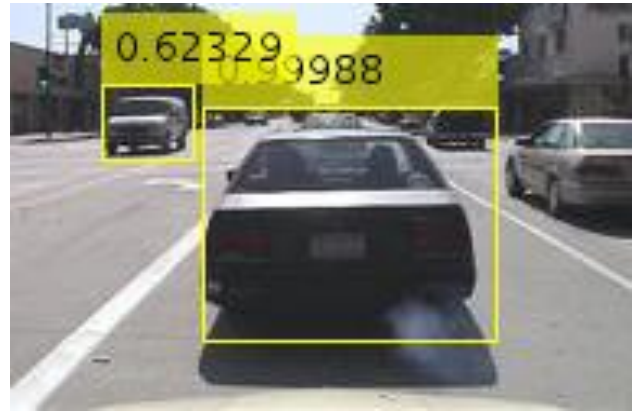

*Figure 1* *Result of Frame for Faster R-CNN on test data set 1 [10]*

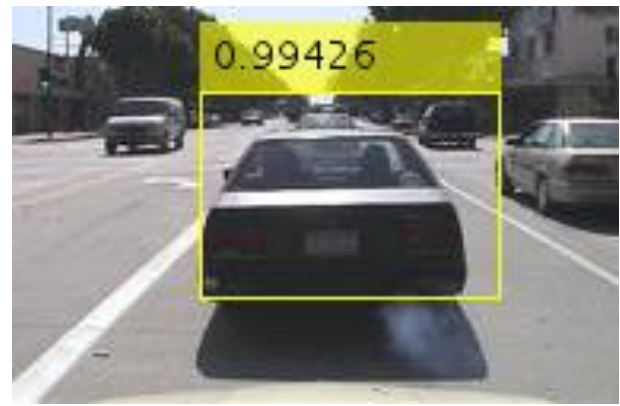

*Figure 2* *Result of Frame for R-CNN on test data set 1 [10]*

Besides, a Precision / Recall (PR) curve is created to highlight the sensitivity state of our detector regarding the degree of recall at various levels, and the mean average precision(map) values are obtained from faster R-CNN and R-CNN respectively, at approximately 0.73, 0.76 and 0.64, 0.65 values. The Precision / Recall (PR) curve is shown in Figure 3 and Figure 7 for fast R-CNN and in Figure 4 and Figure 8 for R-CNN. Furthermore, according to the results, that is shown in Table 1, obtained for the purpose of vehicle detection, results obtained via Faster R-CNN method have higher detection quality and average sensitivity value (mAP) than the results obtained via R-CNN method. In addition, it has been observed that object detection is faster and more reliable via Faster R-CNN method.

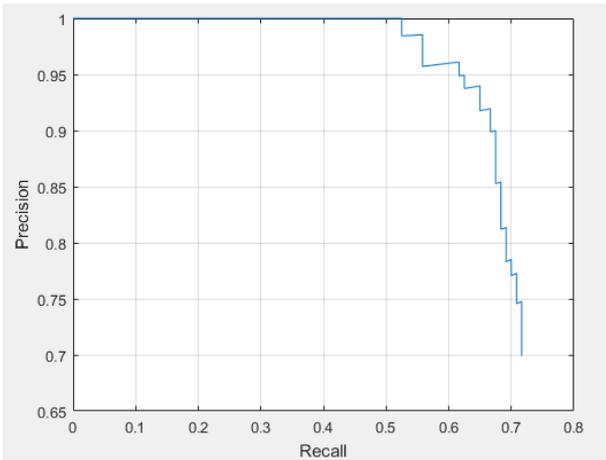

*Figure 3* Precision/Recall (PR) curve for Faster R-CNN on data set 1[10]

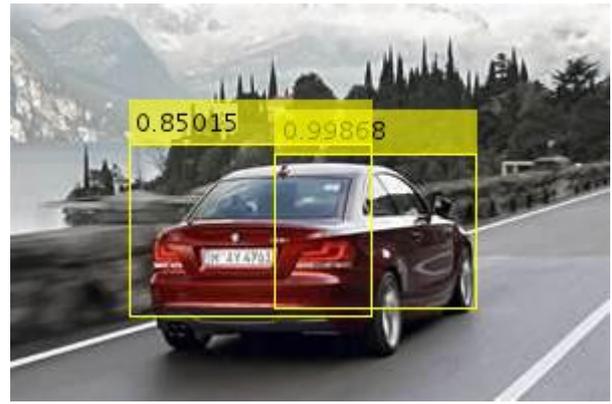

*Figure 6* Result of Frame for R-CNN on test data set 2 [12]

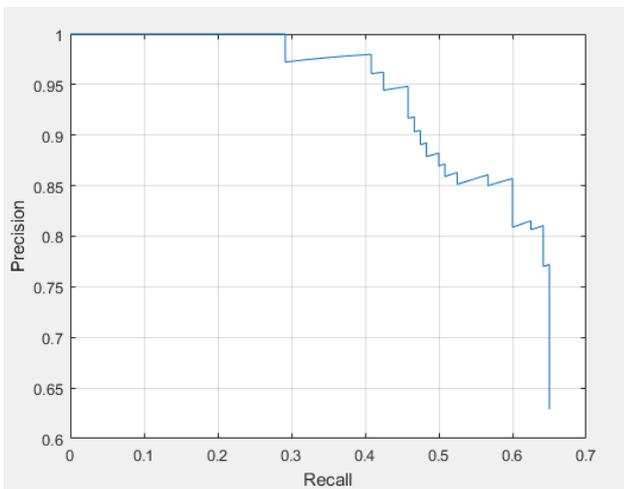

*Figure 4* Precision/Recall (PR) curve for R-CNN on data set 1[10]

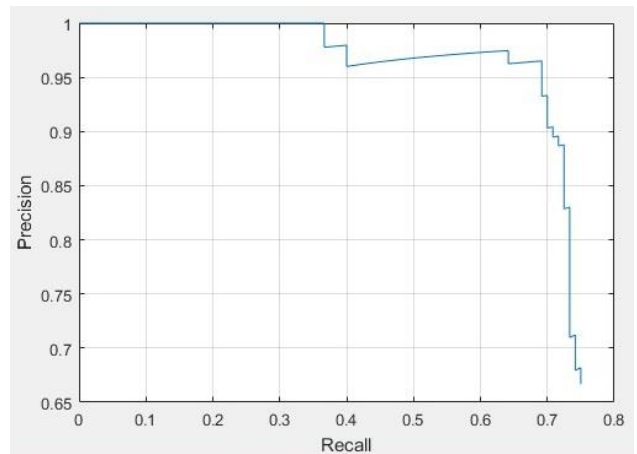

*Figure 7* Precision/Recall (PR) curve for Faster R-CNN on data set 2 [12]

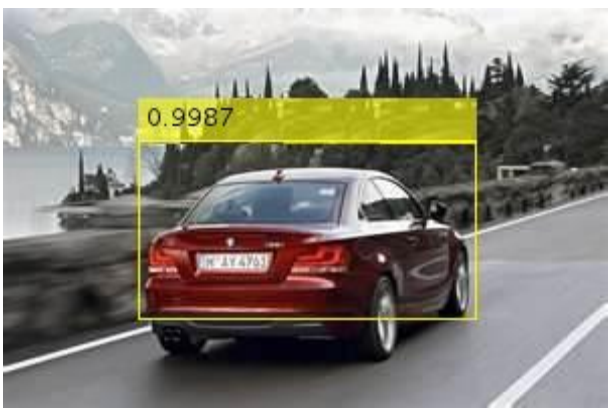

*Figure 5* Result of Frame for Faster R-CNN on test data set 2 [12]

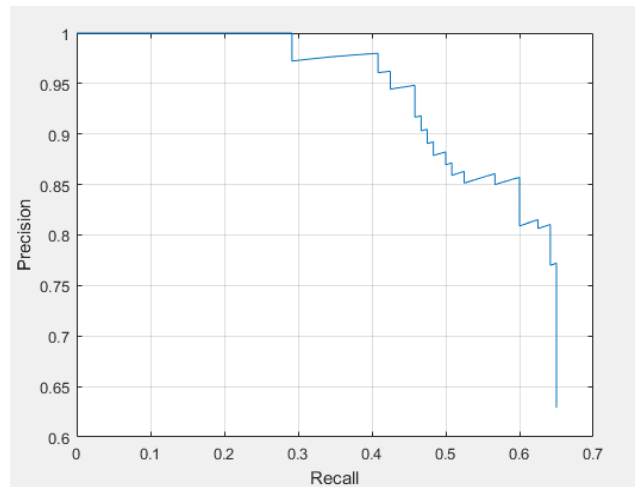

*Figure 8* Precision/Recall (PR) curve for R-CNN on data set 2 [12]

*Table 1  Comparision Table for Faster R-CNN and R-CNN Methods*

| Method | Data set | mAP(%) | test time(h) |
|---|---|---|---|
| *Faster R-CNN* | *Data set 1 [10]* | *72.8* | *0.59* |
| *Faster R-CNN* | *Data set 2 [12]* | *75.7* | *0.74* |
| *R-CNN* | *Data set 1 [10]* | *64.7* | *5.25* |
| *R-CNN* | *Data set 2 [12]* | *65.3* | *6.67* |